\input psfig
\newcommand{\fig}[3]{
   \begin{figure}
      \hfil
      \psfig{figure=#1,height=#2}
      \hfil
      \caption{#3}
   \end{figure}
 }
\documentstyle[aclap]{article}

\title{\vspace{-0.5in}Multilingual phonological analysis and speech synthesis}
\author{John Coleman\thanks{\hspace{0.1in}Oxford
University Phonetics Laboratory/OUPL,
41 Wellington Sq, Oxford OX1 2 JF, UK. Email:
{\tt John. Coleman@Phonetics.Oxford.ac.uk}}\And
Arthur Dirksen\thanks{\hspace{0.1in}Institute for Perception Research/IPO,
P.O. Box 513, 5600 MB Eindhoven, Netherlands. Email:
{\tt adirksen@natlab.research.philips.com}.
The research of Arthur Dirksen has been made possible by a fellowship
of the Royal Netherlands Academy of Arts and Sciences.}\And
Sarmad Hussain\thanks{\hspace{0.1in}Northwestern University. Email:
{\tt shu829@lulu. acns.nwu.edu}}\And
Juliette Waals\thanks{\hspace{0.1in}Research Institute for Language and
Speech/OTS,
Utrecht University, Trans 10, 3512 JK Utrecht, Netherlands. Email:
{\tt Juliette.Waals@let.ruu.nl}}
}

\begin{document}
\bibliographystyle{fullname}
\maketitle
\vspace{-0.5in}
\begin{abstract}
We give an overview of multilingual speech synthesis using
the IPOX system.  The first part discusses work in progress for various
languages: Tashlhit Berber, Urdu and Dutch. The second part discusses a
multilingual phonological grammar, which can be adapted to a particular
language by setting parameters and adding language-specific details.
\end{abstract}

\section{Introduction}
The goal of our research into multilingual speech synthesis is to
maximize the reuse of linguistic rules and data, not just the reuse of
tools such as synthesizer and rule compiler. The reuse of linguistic
rules is facilitated by a declarative, abstract rule notation, such as
constraint-based grammar. In a constraint-based framework, all
constraints must be satisfied conjunctively, and as a result each
constraint stands on its own: it is either a universal property of all
languages, a property of a class of languages, or a language-specific
statement.

In this paper, we give an overview of multilingual work using the
experimental, ``all-prosodic" IPOX synthesis system, based on rules
developed for English \cite{Dir+Cole}. In this
system, input strings are analyzed using declarative, constraint-based
phrase structure grammars, which can be developed using a separate rule
compiler. The representation thus obtained, a metrical-prosodic tree,
is assigned a compositional phonetic interpretation in terms of
parameters for a formant synthesizer. Phonetic interpretation of
syllables is sensitive to metrical structure in that weak nodes are
overlaid on their strong sister constituents.

Section 2 discusses work in progress for various languages and topics:
syllable structure and syllabification in Berber (2.1), prosodic
structure and phonetic interpretation in Urdu (2.2), and temporal
interpretation of syllables in Dutch (2.3). In each case, we will see a
significant amount of reuse of grammars which were developed for
British English, as well as interesting differences.

Section 3 discusses preliminary results of a more ambitious attempt to
develop a multilingual grammar for IPOX, which can be adapted to a
particular language by setting a number of parameters and adding
language-specific details.

\section{Language-specific research}
This section discusses the use of IPOX for various languages. For each
of these languages more work is needed to obtain a full system.

\subsection{Syllable structure and syllabification in Tashlhit Berber}
Syllabification in Tashlhit Berber is challenging for any theory of
syllabification because, according to Dell and Elmedlaoui (1985),
Tashlhit has many syllabic consonants, and numerous consonant-only
words. For example (capital letters denote syllabics):

\begin{verbatim}
(1) tFtKtStt     you sprained it
    tSkRt        you did
    tXzNt        you stored
\end{verbatim}
%%%removed:    tLdi         she pulls

Coleman (forthcoming) presents an analysis of Tashlhit syllable
structure in which phonetic syllabic consonants are phonologically
analyzed as a coproduced vowel and consonant, as in our treatment of
syllabic consonants in English. According to this view, as a vowel is
shortened, the duration between the consonants is reduced, until a
point is reached at which the coda consonant begins as soon as the
onset consonant is released.

In the case of Tashlhit, a disyllabic consonant sequence such as
[tXzNt] is analyzed as having vocalic nuclei in its phonological
representation e.g.  /t@xz@nt/. The internal structure of syllables is
defined by the following phrase structure rules:

\begin{verbatim}
(2) Syl --> (Onset / Rime)
    Rime --> (Nucleus \ Coda)
    Onset --> X:[+cons]
    Nucleus --> X:[-cons]
    Coda --> X:[+cons]
    Coda --> (X:[+cons] \ X:[+cons]).
\end{verbatim}

In these rules, the slash indicates metrical prominence.  The rules are
further constrained in a number of ways. Because the input strings do
not usually contain overt schwas --- these are to be predicted on the
basis of the computed syllable parses --- the empty string and schwa are
both listed in the segment inventory, with the same features. Either
symbol, in nuclear position, indicates that one of the neighbouring
consonants is syllabic. The empty string may also occur in onset or
coda position, in which case it is not parsed as a schwa, but as an
empty consonant. In branching codas the two X's must be filled, and a
sonority constraint applies. Also, onset, nucleus and coda may each be
empty, but we need to ensure that they are not all empty. Consequently,
the grammar includes a constraint prohibiting empty syllables.

CCV words (e.g. /bdu/) are analysed as C@CV, as there is often an
epenthetic schwa between the two consonants.

In this grammar, there are only two ways in which geminates may be
parsed:  tautosyllabically, in the coda, or transsyllabically, in the
coda of one syllable and in the following onset. Word initial geminates
as in e.g. /ttggwa/ will therefore always have a syllabic beginning,
i.e. [CC...] must be analyzed as either /@CC.../ or possibly /C@C.../.
Initial geminates are syllabic, a fact which has no explanation if
syllables may have branching onsets.  The potential ambiguity in forms
like /atta/ is resolved by a constraint that the onset may not be empty
if the previous coda is filled.

A list of 589 Tashlhit words was parsed, and the analyses were checked
by a native speaker regarding the syllable count, placement of syllable
boundaries, and the distribution of schwas (interpreted as syllabicity
of neighbouring consonants). The parses have the right number of
syllables per word in 98\% of cases. In c. 5\% of the test set, the
informant was unsure of his own judgement as to the placement of
syllable boundaries or syllable count.

\subsection{Prosodic structure and phonetic interpretation in Urdu}
Being distantly related to English and Dutch, the phonological grammar
of Urdu is rather similar in some respects. In particular, Urdu has a
right-headed quantity-sensitive metrical structure, causing primary
stress to fall on one of the last three syllables of the word. In order
to compute the weight of syllables, however, we parse them into one,
two or three moras, illustrating the use of IPOX for the implementation
of an alternative theory of syllable structure. The segment inventory
and many details of phonetic interpretation are also somewhat
different:  vowels may be distinctively nasalized, represented e.g.
/a\verb^~^/; /t, d/ are dental; /x/ and /G/ denote voiceless and voiced velar
fricatives, similar to Dutch; /r/ is an alveolar trill and /R/ an
alveolar flap. All stops and affricates, both voiced and voiceless, are
distinctively unaspirated (unmarked) or aspirated (marked with /h/,
e.g. /bh, dh, Ch, Jh/).

As in contemporary metrical analyses of Hindi stress, syllables are
classified as L(ight, one mora), H(eavy, two moras) or S(uperheavy,
three moras). A mora is a V or CV unit or, syllable-finally, a C; thus,
every vowel and syllable-final consonant adds to syllable weight. The
range of syllable patterns, with number of moras, is:  CV (1 mora), V
(1), CV.C (2), V.C (2), CV.V (2), CV.V.C (3), V.V.C (3).

The Maximal Onset Principle is observed in syllable sequences i.e.
...VCV... is parsed by our rules as ...V.CV... If the onset of a non-initial
syllable is otherwise empty, a glottal stop must be inserted in the
input string.

Stress is determined by the following principle:\\

\noindent
(3) The heaviest of the last three syllables of the word bears primary
stress, marked e.g. /mus.ta.'fiiz/ (HL'S).\\

If the heaviest syllable of the word is not among the last three
syllables, it does not receive the main stress, e.g. /kaan.vo.'kee.San/
(SL'HH). If there is a tie between two heavy final syllables, the
penultimate syllable is stressed, e.g. /'aa.paa/, /vaa.'kaa.lat/.
Likewise, in SLH or HLH words, the antepenultimate rather than the
final H syllable bears stress; e.g.  /'C@@d.ha.rii/, /'haa.zi.mah/.

These phenomena can be understood in terms of (3) if we count final
heavy syllables as light, by making the last mora of the last syllable
extrametrical.  This makes final heavy syllables behave like light
syllables (i.e. unstressed), and final superheavy syllables like heavy
syllables (i.e. stressed), e.g.  /ib.raa.'hii(m)/, /vaa.hii.'yaa(t)/.

Figure 1 illustrates the structure assigned to
``mozaavalat" (L'HLH).

\fig{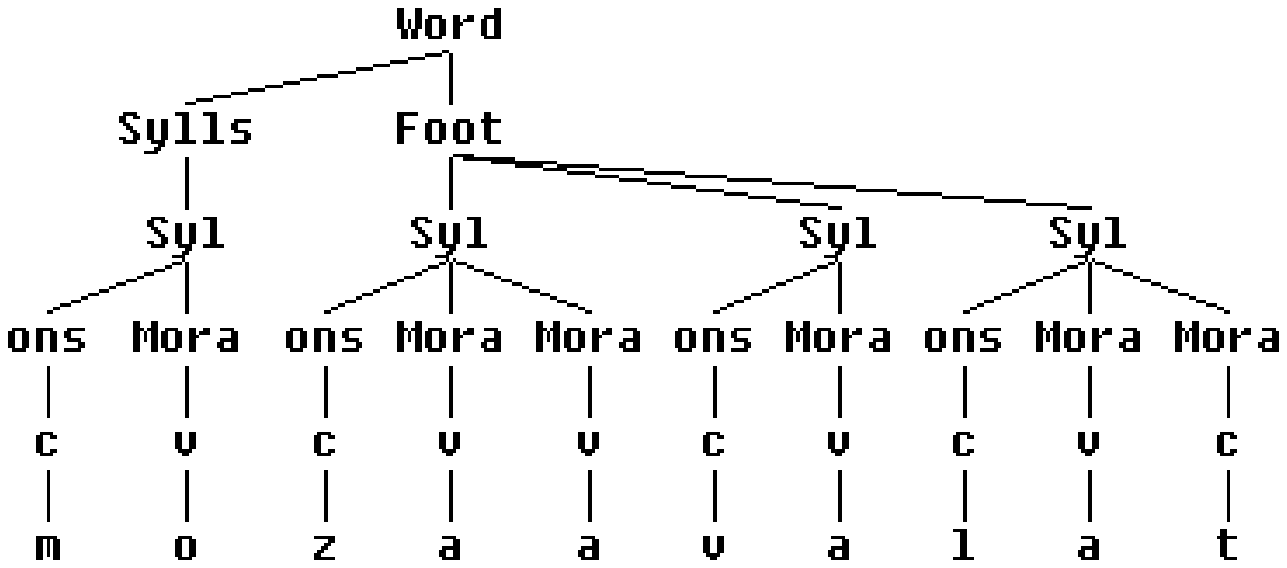}{86pt}{Analysis of Urdu ``mozaavalat"}

In apparent word-final four-mora syllables, as in /bar.xaast/, the
final C is treated as extrametrical, an analysis which is extended to
all final consonant clusters e.g. /taxt/, /Sajr/, /arz/. Syllable
patterns like \verb$CVCC#$ and \verb$CVVCC#$ are possible, but are no heavier
than
three moras. We analyse such syllables as having two [+coda] moras. In
some such cases, the second consonant is syllabic: this is not
currently expressed in our analysis.

Phonetic interpretation of syllables composed of onset and moras
parallels that of syllables composed of onset, rime, nucleus and coda.
C-to-V transitions are modelled by overlaying ons, the nonhead daughter
of the Syl node, on the first Mora of the syllable. V-to-C transitions
are modelled either within a heavy syllable as overlaying a final
C-mora on an internal V-mora, or by overlapping the beginning of one
syllable over the end of the preceding syllable, so that the onset of
the later syllable overlaps the last (C)V-mora of the earlier
syllable.

Many durations and parameter values were based on the IPOX-English
values to begin with, with the addition of measurements of Urdu
acoustics from Hussain (1993). This meant that where Urdu values are as
yet unknown, the system is nevertheless able to produce speech-like
output, albeit with an English accent in some respects. A preliminary
set of durations of long and short vowels in open and closed syllables
with voiced and voiceless onsets was measured from a small database of
monosyllabic words recorded in a sentence frame.  Vowel parameters were
also estimated from this database.

Parameters for English /h/ were used for aspirate /h/. Voiced aspirates
were modelled on English voiceless aspirates, with an earlier onset of
voicing.  Aspiration parameters were added to English affricates to model
Urdu aspirated affricates, the aspiration of voiceless affricates
starting relatively later.

Parameters of Urdu alveolar stops were modelled on English alveolar
stops.  Parameters for Urdu dental stops were collected from our data.
Velar fricatives /x/ and /G/ were modelled on lip-rounded varities of
English /S/ (as in pressure) and /Z/ (as in pleasure), modified to
sound velar.

\subsection{Temporal interpretation of syllables in Dutch}
Waals (forthcoming) reports effects of phonotactic structure on segment
durations in Dutch syllables, specifically onsets. We measured
durations of onset constituents in 151 monosyllabic words embedded in a
carrier phrase in post-focal position, spoken by a male speaker of
Dutch, with two repetitions. We found that segment durations are
reliably predicted by:

1. {\em voicing:} In simple onsets, the ratio between voiced and voiceless
obstruent durations is about 120ms/150ms = 0.8. The same ratio was
found for the total duration of binary onsets: 160ms/200ms = 0.8,
depending on whether the first consonant (an obstruent) is voiced or
not;

2. {\em sonority:} In simple onsets, liquids are shorter than nasals (95ms $<$
110ms), which in turn are shorter than obstruents (120ms/150ms). A
similar effect was found for the second consonant of binary onsets
(50ms $<$ 65ms $<$ 95ms);

3. {\em slot filling:} In binary onsets, the first consonant simply fills
the space that is left. For example, in /sp/, /sm/ and /sl/ the total
duration is 200ms, as predicted by 1, but /s/ is shorter in /sp/ than
in /sm/ and /sl/, as predicted by 2 (200ms--95ms $<$ 200ms--65ms $<$
200ms--50ms);

4. {\em compositionality:} The temporal structure of complex onsets such as
/spl/ follows from principles of compositionality if we assume a
left-branching structure as in (4) below. That is, the duration and
internal temporal structure of /sp/ in /spl/ is the same as in the case
of a binary onset (200ms total, with 95ms for /p/). Also, the duration
of /l/ in /spl/ is the same as in any other cluster with /l/, 50ms.
Thus, we obtain a total duration of 250ms for /spl/, which is correct.

\begin{figure*}
%% DO NOT CHANGE THE SPACING HERE!!!
\begin{verbatim}
     (4)    /  \                   (5)         |----------------l--|  (200ms
for /l/)
           w    s                         |--------------p----|       (200ms
for /p/)
          / \                             |-----s----|
         w   s
         s   p  l                         ^---105----^---95---^-50-^
 (segments)
                                          ^----------250-----------^  (total
onset)
\end{verbatim}
%%%\caption{Metrical and temporal structure of Dutch onset /spl/}
\end{figure*}

These four effects have found a straightforward encoding as rules for
temporal interpretation in IPOX, without loss of generalization. We
will briefly illustrate this for /spl/. The metrical structure assigned
by the parser (reusing the rules for English onset) is shown in (4).
Temporal interpretation in IPOX is done by solving two sets of
constraints, duration constraints and constraints determining the
degree of (non-)overlap between constituents.  Within a syllable, the
total duration of a constituent is assigned to the head (the strong
node) of that constituent, and weak nodes are (fully or partially)
overlaid on their strong sister nodes. Thus, as illustrated in (5),
/l/, being the head of a branching constituent with a voiceless
obstruent, is assigned a duration of 200ms. Since /p/ is also the head
of a branching constituent, it is also assigned a duration of 200ms.
The observed duration of /l/ in clusters, 50ms, is brought into the
equation as the amount of non-overlap with /sp/, so that /sp/ ends 50ms
before /spl/ ends. For /p/ in clusters, we specify 95ms of non-overlap
with the preceding segment. Since no inherent duration is specified for
/s/, it fills the remaining 105ms. The latter follows from a convention
built into IPOX, that is, ``slot filling" is automatic.

\section{Parameterized multilingual speech synthesis}
To complement and extend our work on grammars for various languages in
IPOX, we have recently undertaken a more ambitious attempt to develop
IPOX grammars with built-in multilinguality. In such grammars, a
distinction is made between:  a universal core, which consists of rules
that (to the best of our knowledge) apply to all human languages; a set
of parameterized rules, which define dimensions along which languages
may differ systematically; and language-specific rules and data, which are
needed in those places where languages differ in ways which do not
derive from general considerations. Ideally, with such a setup, a
language is generated by setting a number of parameters and adding
language-specific constraints. However, a long-term research investment is
required to develop the universal, parameterized core and the
language-specific extensions.

To support parameterization of grammars, IPOX includes a facility for
conditional compilation, which is very similar to what is found in many
programming languages. Parameter settings and language-specific rules
are included from files which are kept in separate directories.

Coleman (1991) has shown that constraint-based grammar provides an
excellent vehicle for implementing a parameterized metrical theory of
word stress. In this section, we extend this work to syllable-internal
structure and the phonology-phonetics interface.

As a simple example of parameter setting, consider the representation
of segments. It is generally assumed that voiceless, unaspirated stops
represent the unmarked case cross-linguistically, and that
distinctively voiced and aspirated stops are marked options. This is
expressed in our grammar by two default parameter settings, VoicedStops
= no and AspiratedStops = no. In order to parse a language which
includes voiced and/or aspirated stops, one must change these defaults.
If AspiratedStops is set to yes, the grammar accepts /ph/ in addition
to /p/ as a terminal (example: Mandarin Chinese). By setting
VoicedStops = yes as well, we add /b/ and /bh/ to our segment inventory
(example:  Urdu, see 2.2). A language like Thai, which includes both
voiced and aspirated stops, but no voiced aspirated stops, would need a
language-specific filter: *[+voi, +spread]. English provides a special
case in that /p/ is aspirated in /pit/, but not /spit/. Since the
aspiration is not distinctive, it is considered a matter of phonetic
interpretation (i.e. parameter settings for English include VoicedStops
= yes, AspiratedStops = no).

A more complex example is the representation of syllable weight. (The
analysis presented here is based directly on Zec, 1995). As discussed in
section 2.2 above, a syllable may dominate one, two or three moras. A
syllable is light if it dominates a single mora (and, optionally,
non-moraic material as well), heavy otherwise. We assume that this is
true universally, but that the case of three moras must be negotiated
by setting a parameter SuperHeavySyllable = yes.  However, languages
differ in the sonority classes accepted in various syllable-internal
positions. For example, English differs from, say, Cairene Arabic in
that the former allows syllabic sonorants, whereas the latter accepts
only vowels in this position. In a similar fashion, some languages
require a mora to dominate a sonorant (or even a vowel), prohibiting
obstruents in this position. In the our grammar, this is implemented by
adding ``syllabicity" and ``moraicity" constraints to syllables and
moras, as follows:

\begin{verbatim}
(6) Syl:[-heavy] --> Mora:[SYLLABIC]
    Syl:[+heavy] -->
                (Mora:[SYLLABIC] \ Mora)
    Mora:[MORAIC] --> X.
\end{verbatim}

\verb^SYLLABIC^ and \verb^MORAIC^ are macros which expand to feature structures
during rule compilation, and the exact content of these feature
structures is parameterized with respect to the sonority classes
allowed in these positions: [--cons] (only vowels), [+son] (only
sonorants) or [ ] (all segments).

By setting four parameters, various types of weight-sensitive languages
are generated. For example, if moras are limited to [+son] material,
/pin/ counts as heavy, but /pit/ is either ruled out, or /t/ is parsed
as an adjunction to the syllable (depending on yet another parameter
setting), making /pit/ a light syllable. A maximally unrestricted
language accepts any segment in syllabic position. According to
standard views Berber is an example (but see 2.1).

One advantage of the approach outlined above is that it allows us to
stan\-dardize the phonology-phonetics interface, thus making reuse of
phonetic interpretation rules much easier. All phonetic interpretation
rules are of the form Prosodic\-Structure \verb^=>^ Phonetic\-Exponents.
Keeping
what appears on the lefthand side of the equation more or less constant
from one language to the other makes it easier to deal with the
righthand side.

Also, it seems possible to extend our approach to phonetic
interpretation. In IPOX, a distinction is made between the general
shape of synthesis parameter tracks and the actual values of parameters
at different points in time. For example, in our English grammar CV
transitions are generated from general descriptions of formant
trajectories such as:

\begin{verbatim}
(7) F2(20%, 50%, 90%, 100%, 100%+F2End)
    = (?, F2Value, F2Value,
       F2Locus+F2Coart*(F2Vowel-F2Locus),
       ?F2Vowel).
\end{verbatim}

Here F2 is a synthesis parameter. The numbers in the lefthand side of
the equation are points in time (expressed as percentages of the
duration of a constituent) at which the corresponding values on the
right are in effect. The variables F2End, F2Value, F2Locus and F2Coart
are evaluated by consulting lookup tables. The use of these lookup
tables allows partial generalizations of the form ``F2Locus is such and such
for all labial obstruents before a front vowel". It seems to us that
these general descriptions can be recast as universals of phonetic
interpretation, and that language-specific differences can be accounted
for in lookup tables. Initial values in lookup tables are obtained by
estimating reasonable cross-linguistic defaults.

\section{Conclusion}
We have shown that standard analyses from the contemporary phonological
literature can be expressed in the IPOX rule formalism with ease, and
given a phonetic interpretation, however approximate initially. We know
of no other system which permits analyses at this level of phonological
sophistication in combination with speech output.

In addition, we have sketched an approach which allows us to do justice
to cross-linguistic generalizations by incorporating a mechanism for
parameterization of grammars.

More information, demos, as well as an evaluation copy of IPOX can be
found at \verb^ftp://chico.phon.ox.ac.uk/pub/ipox/^ .

\end{document}